\documentclass[11pt]{article}
 
\PassOptionsToPackage{numbers, compress}{natbib}
\usepackage[utf8]{inputenc}
\usepackage{amsmath}
\usepackage{amsfonts}
\usepackage{mathrsfs}
\usepackage{amssymb}
\usepackage{amsthm}

\usepackage{libertine}
\usepackage{wrapfig}
\usepackage{graphicx}
\usepackage{multirow}
\usepackage{epstopdf}
\usepackage{multicol}
\usepackage{subfigure}
\usepackage{booktabs}

\usepackage{algorithmic}
\usepackage{algorithm}
\usepackage[algo2e,linesnumbered]{algorithm2e}
\usepackage{stackengine}
\usepackage{graphicx}
\usepackage{multirow}
\usepackage{epstopdf}
\usepackage{multicol}
\usepackage{caption}
\usepackage{subfigure}
\usepackage{enumitem}
\usepackage{amsmath}
\usepackage{pifont}
\usepackage{natbib}
\usepackage{bbm}
\usepackage{makecell}
\usepackage{mathtools}
\usepackage[table]{xcolor}
\definecolor{mydarkred}{rgb}{0.6,0,0}
\definecolor{mydarkgreen}{rgb}{0,0.6,0}
\usepackage[colorlinks,
linkcolor=mydarkred,
citecolor=mydarkgreen]{hyperref}
\usepackage{url}
\usepackage{bm}
\usepackage{pifont}
\usepackage{stfloats}
\usepackage[T1]{fontenc}
\usepackage{arydshln}
\usepackage{colortbl}
\usepackage{balance}

\usepackage{tikz}
\newcommand{\yn}{\tilde{y}}
\newcommand{\Yn}{\tilde{Y}}
\newcommand{\px}{p_{\theta_1}(x|y,z)}
\newcommand{\pyn}{p_{\theta_2}(\tilde{y}|y, x)}
\newcommand{\Px}{p_{\theta_1}(X|Y,Z)}
\newcommand{\Pyn}{p_{\theta_2}(\tilde{Y}|Y, X)}

\newcommand{\qz}{q_{\phi_2}(z|y,x)}
\newcommand{\qy}{q_{\phi_1}(y|x)}
\newcommand{\Qz}{q_{\phi_2}(Z|Y,X)}
\newcommand{\Qy}{q_{\phi_1}(Y|X)}

\newcommand{\q}{q_{\phi}(z,y|x)}

\newcommand\independent{\protect\mathpalette{\protect\independenT}{\perp}}

\makeatletter
\renewcommand*{\@fnsymbol}[1]{\ensuremath{\ifcase#1\or
\dagger\or \ddagger\or
    \mathsection\or \mathparagraph\or \|\or **\or \dagger\dagger
    \or \ddagger\ddagger \else\@ctrerr\fi}}
\makeatother
\def\independenT#1#2{\mathrel{\rlap{$#1#2$}\mkern2mu{#1#2}}}

\usetikzlibrary{positioning,shapes,arrows}
\title{Instance-dependent Label-noise Learning under a Structural Causal Model}

\author{
  Yu Yao$^{1}$,
  Tongliang Liu$^{1}\thanks{Correspondence to Tongliang Liu (tongliang.liu@sydney.edu.au).}$,
  Mingming Gong$^{2}$, \\
  {  Bo Han$^{3}$, Gang Niu$^4$}, Kun Zhang$^5$\\[1ex]
  $^1$University of Sydney;
  $^2$University of Melbourne;
  $^3$Hong Kong Baptist University;\\
  $^4$RIKEN AIP;
  $^5$Carnegie Mellon University\\
}

\date{}
\begin{document}

\maketitle

\begin{abstract}
Label noise will degenerate the performance of deep learning algorithms because deep neural networks easily overfit label errors. 
Let $X$ and $Y$ denote the instance and clean label, respectively. 
When $Y$ is a cause of $X$, according to which many datasets have been constructed, e.g., \textit{SVHN} and \textit{CIFAR}, the distributions of $P(X)$ and $P(Y|X)$ are entangled. This means that the unsupervised instances are helpful to learn the classifier and thus reduce the side effect of label noise. However, it remains elusive on how to exploit the causal information to handle the label noise problem.
In this paper, by leveraging a structural causal model, we propose a novel generative approach for instance-dependent label-noise learning. In particular, we show that properly modelling the instances will contribute to the identifiability of the label noise transition matrix and thus lead to a better classifier. 
Empirically, our method outperforms all state-of-the-art methods on both synthetic and real-world label-noise datasets.
\end{abstract}

\newpage
\section{Introduction}
% ***Paragraph 1*** introduce the importance of the label noise problem:
% \begin{enumerate}
%     \item history of label-noise learning
%     \item label noise widely exist in big data. 
%     \item the impact of label noise
%     \item practical value of label-noise learning
% \end{enumerate}

Learning with noisy labels can be dated back to \citep{angluin1988learning}, which has recently drawn a lot of attention \citep{liu2021importance,cheng2020learning,xia2019anchor,zhu2020second}.
In real life, large-scale datasets are likely to contain label noise. It is mainly because that many cheap but imperfect data collection methods such as crowd-sourcing and web crawling are widely used to build large-scale datasets. Training with such data can lead to poor generalization abilities of deep neural networks because they can memorize noisy labels \citep{arpit2017closer,yao2020searching}.

To improve the generalization ability of learning models training with noisy labels, one family of existing label-noise learning methods is to model the label noise \citep{liu2016classification,natarajan2013learning,zhu2021clusterability,patrini2017making,yao2020dual}. Specifically, these methods try to reveal the transition relationship from clean labels to noisy labels of instances, i.e., the distribution $P(\Yn|Y,X)$, where $\Yn$, $Y$ and $X$ denote the random variable for the noisy label, latent clean label and instance, respectively. The advantage of modelling label noise is that given only the noisy data, when the transition relationship is identifiable, classifiers can be learned to converge to the optimal ones defined by the clean data, with theoretical guarantees. 
However, the transition relationship is not identifiable in general. To make it identifiable, different assumptions have been made on the transition relationship. For example, Natarajan et al. \citep{natarajan2013learning} assumes that the transition relationship is instance independent, i.e.,  $P(\Yn|Y,X)=P(\Yn|Y)$; Xia et al. \citep{xia2020part} assumes that $P(\Yn|Y,X)$ is dependent on different parts of an instance. Cheng et al. \citep{cheng2020learning} assumes that the label noise rates are upper bounded. In practice, these assumptions may not be satisfied and hard to be verified given noisy data alone.

Inspired by causal learning \citep{pearl2009causality,peters2017elements,scholkopf2019causality}, we provide a new causal perspective of label-noise learning named \textit{CausalNL} by exploiting the causal information to further contribute to the identifiability of the transition matrix $P(\Yn | Y,X)$ other than making assumptions directly on the transition relationship. Specifically, we assume that the data containing instance-dependent label noise is generated according to the causal graph in Fig.~\ref{fig:casual}. In real-world applications, many datasets are generated according to the proposed generative process. For example, for the Street View House Numbers (SVHN) dataset \citep{netzer2011svhn}, $X$ represents the image containing the digit; $Y$ represents the clean label of the digit shown on the plate; 
$Z$ represents the latent variable that captures the information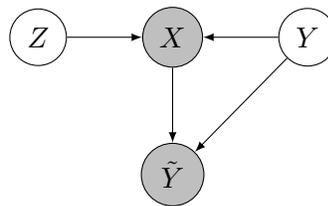
\begin{wrapfigure}{h}{5cm}
    \centering
    \begin{tikzpicture}[
      mynode/.style={draw,circle,align=center}
    ]
    \node[mynode, fill=black!25] (sp) {$X$};
    \node[mynode, left =of sp] (zi) {$Z$};
    \node[mynode,below =of sp, fill=black!25] (gw) {$\Yn$};
    \node[mynode, right=of sp] (ra) {$Y$};
    \path (ra) edge[-latex] (sp)
    (sp) edge[latex-] (zi) 
    (sp) edge[-latex] (gw) 
    (gw) edge[latex-] (ra);
    \end{tikzpicture} 
    
\caption{A graphical causal model which reveals a generative process of the data which contains instance-dependent label noise, where the shaded variables are observable and the unshaded variables are latent.}\label{fig:casual}
\end{wrapfigure} affecting the generation of the images, e.g., orientation, lighting, and font style. In this case, $Y$ is clearly a cause of $X$ because the causal generative process can be described in the following way. First, the house plate is generated according to the street number and attached to the front door. Then, the house plate is captured by a camera (installed in a Google street view car) to form $X$, taking into account of other factors such as illumination and viewpoint. Finally, the images containing house numbers are collected and relabeled to form the dataset. 
% Note that the images are initially generated by given a range of digit numbers. For example, if your house number is "66", you may write the number on a plate and attach it to your front door. It is clear that $Y$ is a cause of $X$. The Google street view images containing house numbers were then collected and relabeled in the dataset. 
Let us denote the annotated label by the noisy label $\Yn$ as the annotator may not be always reliable, especially when the dataset is very large but the budget is limited. During the annotation process, the noisy labels were generated according to both the images and the range of predefined digit numbers. Hence, both $X$ and $Y$ are causes of $\Yn$. Note that most existing image datasets are collected with the causal relationship that $Y$ causes $X$, e.g., the widely used \textit{FashionMNIST} and \textit{CIFAR}. When we synthesize instance-dependent label noise based on them, we will have the causal graph illustrated in Fig.~\ref{fig:casual}. Note also that some datasets are generated with the causal relationship that $X$ causes $Y$. Other than using domain knowledge, the different causal relationships can be verified by employing causal discovery \citep{spirtes2000causation,spirtes2016causal,peters2017elements}.

\begin{figure}[t]
\begin{center}
\subfigure[]{
    \label{fig:moona}
    \includegraphics[width=0.24\linewidth]{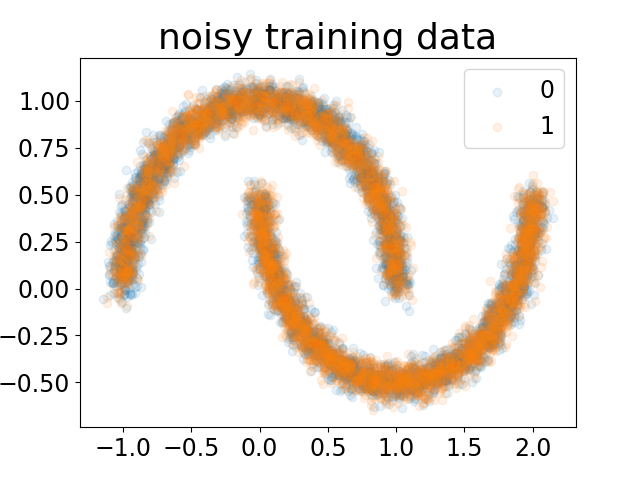}
}
\hspace{-10px}
\subfigure[]{
    \label{fig:moonb}
    \includegraphics[width=0.24\linewidth]{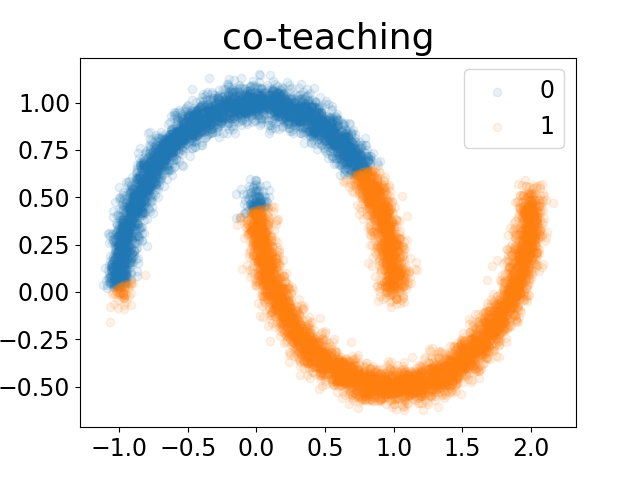}
}
\hspace{-10px}
\subfigure[]{
    \label{fig:moonc}
    \includegraphics[width=0.24\linewidth]{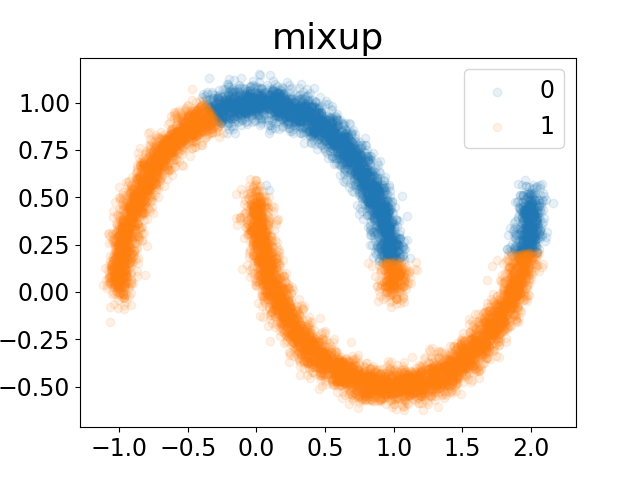}
}
\hspace{-10px}
\subfigure[]{
    \label{fig:moond}
    \includegraphics[width=0.24\linewidth]{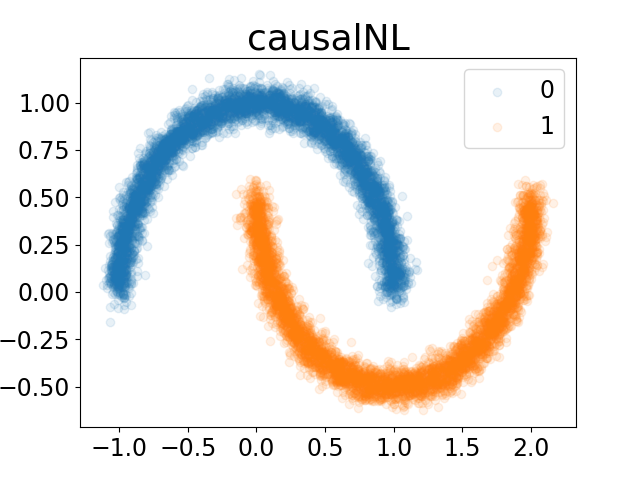}}
    
\end{center}
\vspace{-10pt}
\caption{(a) An illustration of the MOON training dataset which contains $45\%$ of instance-dependent label noise. (b)-(d) The illustration of the classification performance of co-teaching, mixup and our method, respectively. }
\label{fig:moon}
\vspace{-5pt}
\end{figure}

When the latent clean label $Y$ is a cause of $X$, $P(X)$ will contain some information about $P(Y|X)$. 
This is because, under such a generative process, the distributions of $P(X)$ and $P(Y|X)$ are entangled \citep{scholkopf2012causal}. To help estimate $P(Y|X)$ with $P(X)$, we make use of the causal generative process to estimate $P(X|Y)$, which directly benefits from $P(X)$ by generative modeling. The modeling of $P(X|Y)$ in turn encourages the identifiability of the transition relationship and benefits the learning $P(Y|X)$.
For example, in Fig.~\ref{fig:moona}, we have added instance-dependent label-noise with rate  45\% (i.e., IDLN-45\%) to the MOON dataset and employed different methods \citep{han2018co, zhang2017mixup} to solve the label-noise learning problem. As illustrated in Fig.~\ref{fig:moonb} and Fig.~\ref{fig:moonc}, previous methods fail to infer clean labels. In contrast, by constraining the conditional distribution of the instances, i.e., restricting the data of each class to be on a manifold by setting the dimension of the latent variable $Z$ to be 1-dimensional, the label transition as well as the clean labels can be successfully recovered (by the proposed method), which is showed in Fig.~\ref{fig:moond}.

% ***Paragraph 5*** Organization of the paper\footnote{tl: reorganise the paragraph. We need to intuitively express our idea. For example, introduce VAE. In the decoder part, A=BC. "B" and "C" are the feature distribution and label noise transition matrix, respectively. We introduce domain knowledge on B will be equivalent to contraining B.  Then link to figure 1.}
% \begin{enumerate}
%     \item Section 2: A Casual View of Label Noise
%     \item Section 3: explain our method 
%     \item Section 4: Experiments on mnist, Fmnist, Cifar10, SVHN, and Cloth1M
%     \item Section 5: conclusion     
% \end{enumerate}

% We need one more paragraph to show the intuition of our proposed method to let the reader believe in our proposed method even without looking at the experiments.

Specifically, to make use of the causal graph to contribute to the identifiability of the transition matrix, we propose a causally inspired deep generative method which models the causal structure with all the observable and latent variables, i.e., the instance $X$, noisy label $\Yn$, latent feature $Z$, and the latent clean label $Y$. The proposed generative model captures the variables' relationship indicated by the causal graph. Furthermore, built on the variational autoencoder (VAE) framework \citep{kingma2013auto}, we build an inference network which could efficiently infer the latent variables $Z$ and $Y$ when maximising the marginal likelihood $p(X,\tilde{Y})$ on the given noisy data. In the decoder phase, the data will be reconstructed by exploiting the conditional distribution of instances $P(X|Y,Z)$ and the transition relationship $P(\Yn |Y,X)$, i.e., $$ p_\theta(X, \Yn) = \int_{z,y} P(Z=z)P(Y=y)p_{\theta_1}(X|Y=y,Z=z)p_{\theta_2}(\Yn|Y=y,X) \mathrm{d}z\mathrm{d}y$$ will be exploited, where $\theta:= (\theta_1,\theta_2)$ are the parameters of the causal generative model (more details can be found in Section \ref{section3}). In a high level, according to the equation, given the noisy data and the distributions of $Z$ and $Y$, constraining $p_{\theta_1}(X|Y,Z)$ will also greatly reduce the uncertainty of $p_{\theta_2}(\Yn|Y,X)$ and thus contribute to the identifiability of the transition matrix. Note that adding a constraint on $p_{\theta_1}(X|Y,Z)$ is natural, for example, images often have a low-dimensional manifold \citep{belkin2006manifold}. We can restrict $P(Z)$ to fulfill the constraint on $p_{\theta_1}(X|Y,Z)$. By exploiting the causal structure and the constraint on instances to better model label noise, the proposed method significantly outperforms the baselines. When the label noise rate is large, the superiority is evidenced by a large gain in the classification performance.

% To let the latent variable $Z$ be the lower-dimensional representations of $X$ and $Y$ to be the estimated clusters, we set the dimension of $Z$ to be much lower than the dimension of $X$, and $Y$ to be a one-hot vector. In such a way, the model has to make the information provided by $Y$ to be useful to reconstruct $X$ well during the parameter optimisation. Suppose that the dataset follows the manifold assumption, i.e., the instances of the same class lays on the same manifold and the instances of the different classes lays on the different manifolds (see Fig.~\ref{fig:moond}), then the latent variable $Y$ could be an indicator represents which manifold an instance $X$ belonging to and $Z$ is the manifold representation. To further link the clusters to clean labels, we could keep the cluster numbers to be consistent with the noisy labels on some high reliable examples. To achieve this, our method is trained in a expectation maximization favor, i.e., the reliable examples are selected dynamically during the update of parameters of our model by using co-training technique [cite].

The rest of the paper is organized as follows. In Section~\ref{section2}, we briefly review the background knowledge of label-noise learning and causality. 
% Then we introduce the generation process of the instance-dependent label noise. 
In Section~\ref{section3}, we formulate our method,  named \textit{CausalNL}, and discuss how it helps to learn a clean classifier, followed by the implementation details. The experimental validations are provided in Section~\ref{section4}. Section~\ref{section5} concludes the paper.

\section{Noisy Labels and Causality} \label{section2}
In this section, we first introduce how to model label noise. Then, we introduce the structural causal model and discuss how to exploit the model to encourage the identifiability of the transition relastionship and help learn the classifier.

\textbf{Transition Relationship} \ \ 
By only employing data with noisy labels to build statistically consistent classifiers, which will converge to the optimal classifiers defined by using clean data, the transition relationship $P(\tilde{Y}|Y,X)$ has to be identified. Given an instance, the conditional distribution can be written in an $C\times C$ matrix which is called the transition matrix \citep{patrini2017making,xia2019revision,xia2020part}, where $C$ represents the number of classes. Specifically, for each instance $x$, there is a transition matrix $T(x)$. The $ij$-th entry of the transition matrix is $T_{ij}(x)=P(\tilde{Y}=j|Y=i, X=x)$ which represents the probability that the instance $x$ with the clean label $Y=i$ will have a noisy label $\tilde{Y}=j$.

The transition matrix has been widely studied to build statistically consistent classifiers, because the clean class posterior distribution $P(\bm{Y}|x)=[P(Y=1|X=x),\ldots, P(Y=C|X=x)]^\top$ can be inferred by using the transition matrix and the noisy class posterior $P(\tilde{\bm{Y}}|x)=[P(\tilde{Y}=1|X=x),\ldots, P(\tilde{Y}=C|X=x)]^\top$, i.e., we have $P(\tilde{\bm{Y}}|x)=T(x)P(\bm{Y}|x)$.
Specifically, the transition matrix has been used to modify loss functions to build risk-consistent estimators, e.g., \citep{goldberger2016training, patrini2017making,yu2018learning,xia2019anchor}, and has been used to correct hypotheses to build classifier-consistent algorithms, e.g., \citep{natarajan2013learning, scott2015rate, patrini2017making}. Moreover, the state-of-the-art statically inconsistent algorithms \citep{jiang2018mentornet, han2018co} also use diagonal entries of the transition matrix to help select reliable examples used for training.

However, in general, the distribution $P(\tilde{Y}|Y,X)$ is not identifiable \citep{xia2019anchor}. To make it identifiable, different assumptions have been made. The most widely used assumption is that given clean label $Y$, the noisy label $Y$ is conditionally independent of instance $X$, i.e., $P(\tilde{Y}|Y,X) = P(\tilde{Y}|Y)$. Under such an assumption, the transition relationship $P(\tilde{Y}|Y)$ can be successfully identified with the anchor point assumption \citep{liu2016classification,yao2020dual,li2021provably}. However, in the real-world scenarios, this assumption can be hard to satisfied.
Although $P(\tilde{Y}|Y)$ can be used to approximate $P(\tilde{Y}|Y,X)$, the approximation error can be too large in many cases. As for the efforts to model the instance-dependent transition matrix directly, existing methods rely on various assumptions, e.g., the bounded noise rate assumption \citep{cheng2017learning}, the part-dependent label noise assumption \citep{xia2020part}, and the requirement of additional information about the transition matrix \citep{berthon2020idn}. Although the assumptions help the methods achieve superior performance empirically, the  assumptions are difficulty to verify or fulfill, limiting their applications in practice.

\textbf{Structural Causal Model} \ \
Motivated by the limitation of the current methods, we provide a new causal perspective to learn the identifiable of instance-dependent label noise. Here we briefly introduce some background knowledge of causality \citep{pearl2009causality} used in this paper. A structural causal model (SCM) consists of a set of variables connected by a set of functions.
It represents a flow of information and reveals the causal relationship among all the variables, providing a fine-grained description of the data generation process. The causal structure encoded by SCMs can be represented as a graphical casual model as shown in Fig.~\ref{fig:casual}, where each node is a variable and each edge is a function. The SCM corresponding to the graph in Fig.~\ref{fig:casual} can be written as
\begin{align}
    &Z=\epsilon_Z,  \ \ Y =\epsilon_Y, \ \   X = f(Z,Y,\epsilon_{X}), \ \  \tilde{Y} = f(X,Y,\epsilon_{\tilde{Y}}),
\end{align}
where $\epsilon_{\cdot}$ are independent exogenous variables following some distributions. The occurrence the exogenous variables makes the generation of $X$ and $\tilde{Y}$ be a stochastic process. Each equation specifies a distribution of a variable conditioned on its parents (could be an empty set).
% It could be equivalent to sampling $X$ and $\tilde{Y}$ from $P(X|Y, Z)$ and $P(\tilde{Y}|Y, X)$, respectively.

By observing the SCM, the helpfulness of the instances to learning the classifier can be clearly explained. Specifically, the instance $X$ is a function of its label $Y$ and latent feature $Z$ which means that the instance $X$ is generated according to $Y$ and $Z$. Therefore $X$ must contains information about its clean label $Y$ and latent feature $Z$. That is the reason that $P(X)$ can help identify $P(Y|X)$ and also $P(Z|X)$. However, since we do not have clean labels, it is hard to fully identify $P(Y|X)$ from $P(X)$ in the unsupervised setting. For example, on the MOON dataset shown in Fig. \ref{fig:moon} , we can possibly discover the two clusters by enforcing the manifold constraint, but it is impossible which class each cluster belongs to. We discuss in the following that we can make use of the property of $P(X|Y)$ to help model label noise, i.e., encourage the identifiability of the transition relationship, and thereby learning a better classifier.

Specifically, 
under the Markov condition \citep{pearl2009causality}, which intuitively means the independence of exogenous variables,
the joint distribution $P(\tilde{Y},X, Y,Z)$ specified by the SCM can be factorized into the following
\begin{equation}
    P(X,\tilde{Y}, Y,Z) = P(Y)P(Z)P(X|Y,Z)P(\tilde{Y}|Y,X).
    \label{eq:joint}
\end{equation}
This motivates us to extend VAE \citep{kingma2013auto} to perform inference in our causal model to fit the noisy data in the next section. In the decoder phase, given the noisy data and the distributions of $Z$ and $Y$, adding a constraint on $P(X|Y,Z)$ will reduce the uncertainty of the distribution $P(\Yn|Y,X)$. In other words, modeling of $P(X|Y,Z)$ will encourage the identifiability of the transition relationship and thus better model label noise. Since $P(\Yn|Y,X)$ functions as a bridge to connect the noisy labels to clean labels, we therefore can better learn $P(Y|X)$ or the classifier by only use the noisy data.
% The advantage of the above factorization is that all these distributions are independent to each other, which is also known as the independent causal mechanisms principle \citep{scholkopf2012causal}. 

There are normally two ways to add constraints on the instances, i.e., assuming a specific parametric generative model or introducing prior knowledge of the instances. In this paper, since we mainly study the image classification problem with noisy labels, we focus on the manifold property of images and add the low-dimensional manifold constraint to the instances.

\begin{figure}[t]
    \centering
    \includegraphics[width=1\textwidth]{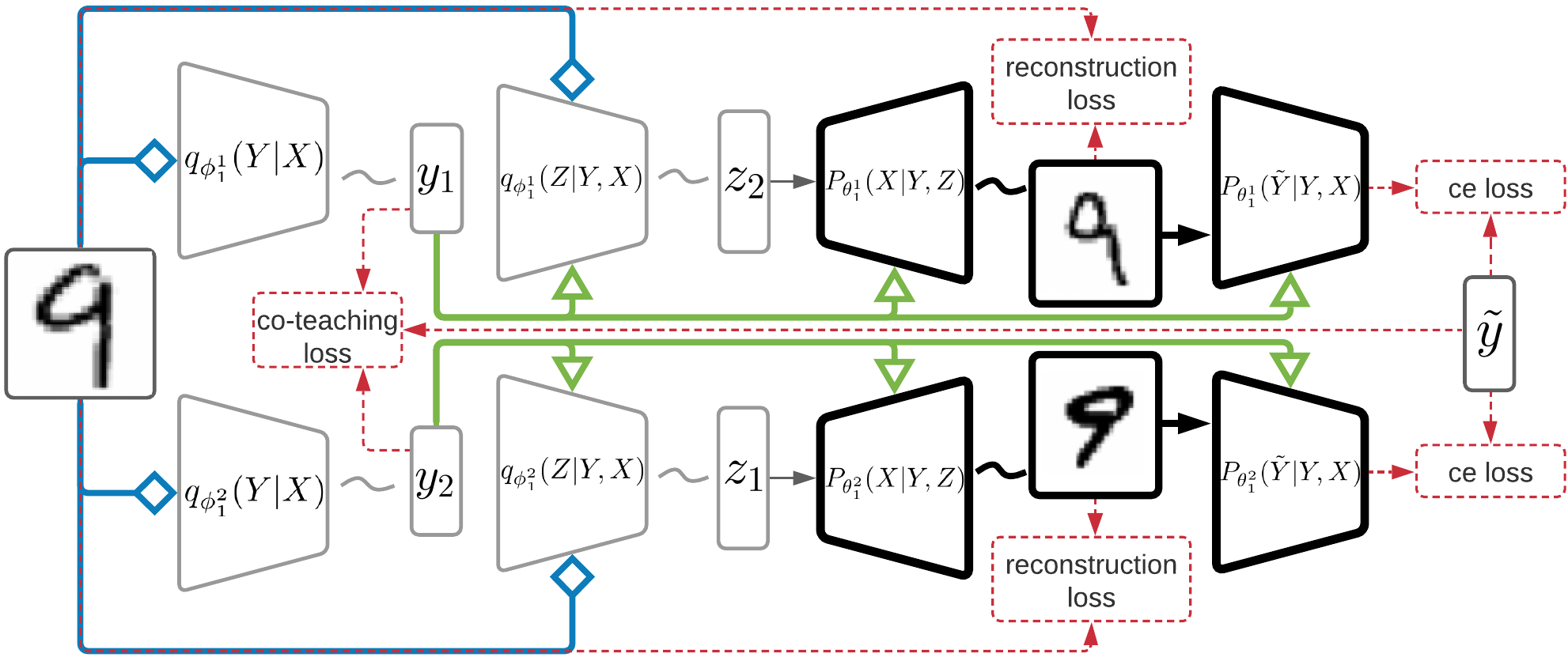}
    \caption{A working flow of our method.}
    \label{fig:wf}
\end{figure}

\section{Causality Captured Instance-dependent Label-Noise Learning}\label{section3}

In this section,  we propose a structural generative method which captures the causal relationship and utilizes $P(X)$ to help identify the label-noise transition matrix, and therefore, our method leads to a better classifier that assigns more accurate labels.

\subsection{Variational Inference under the Structural Causal Model}

To model the generation process of noisy data and to approximate the distribution of the noisy data, our method is designed to follow the causal factorization (see Eq.~\eqref{eq:joint}). Specifically, our model contains two decoder networks which jointly model a distribution $ p_\theta( X, \Yn | Y, Z)$ and two encoder (inference) networks which jointly model the posterior distribution $q_{\phi}(Z,Y|X)$. Here we discuss each component of our model in detail.

Let the two decoder networks model the distributions $\Px$ and $\Pyn$, respectively. Let $\theta_1$ and $\theta_2$ be learnable parameters of the distributions. Without loss of generality, we set $p(Z)$ to be a standard normal distribution and $p(Y)$ be a uniform distribution. Then, modeling the joint distribution in Eq.~\eqref{eq:joint} boils down to modeling the distribution $p_\theta(X, \Yn| Y, Z)$, which is decomposed as follows:
\begin{equation}\label{decoder}
 p_\theta(X, \Yn|Y, Z) = \Px\Pyn.
\end{equation}
To infer latent variables $Z$ and $Y$ with only observable variables $X$ and $\Yn$, we could design an inference network which model the variational distribution $q_{\phi}(Z,Y|\Yn,X)$. Specifically, let $\Qz$ and $q_{\phi_1}(Y|\Yn,X)$ be the distributions parameterized by learnable parameters $\phi_1$ and $\phi_2$, the posterior distribution can be decomposed as follows: 
\begin{equation}\label{encoder_o}
    q_{\phi}(Z,Y|\Yn,X) = \Qz q_{\phi_1}(Y|\Yn,X),
\end{equation}
where we do not include $\Yn$ as a conditioning variable in $\Qz$ because the causal graph implies $Z \independent \Yn|X,Y$. One problem with this posterior form is that we cannot directly employ $q_{\phi_1}(Y|\Yn,X)$ to predict labels on the test data, on which $\Yn$ is absent.  

% It is worth to mention that, to calculate the posterior probability for a test data point $x$ being the label $y$, Bayes' rule [CITES] and Monte Carlo approximation [CITES] are usually employed by generative models, i.e., to approximate $P_{\theta_1}(y|x) = p(y)\int_{z} \px P(z) \de{z} \text{/} p(x)$. The disadvantage is that both the the integration and marginal distribution $p(x)$ is hard to be accurately approximated, especially when both the data $x$ and the latent feature $z$ have high dimension, i.e., curse of dimensionality [CITES]. The large estimation errors could lead to inaccurate estimation of the posterior probability. 
To allow our method efficiently and accurately infer clean labels, we approximate $q_{\phi_1}(Y|\Yn,X)$ by assuming that given the instance $X$, the clean label $Y$ is conditionally independent from the noisy label $\Yn$, i.e., $q_{\phi_1}(Y|\Yn,X)= q_{\phi_1}(Y|X)$. This approximation does not have very large approximation error because the images contain sufficient information to predict the clean labels. Thus, we could simplify Eq.~\eqref{encoder_o} as follows
\begin{equation} \label{encoder}
    q_{\phi}(Z,Y|X) = \Qz \Qy,
\end{equation}
such that our encoder networks model $\Qz$ and $\Qy$, respectively. In such a way, $\Qy$ can be used to infer clean labels efficiently. We also found that the encoder network modelling $\Qy$ acts as a regularizer which helps to identify $\Pyn$. Moreover, be benefited from this, our method can be a general framework which can easily integrate with the current discriminative label-noise methods \citep{xia2019anchor,malach2017decoupling,han2018co}, and we will showcase it by collaborating co-teaching \citep{han2018co} with our method.

\textbf{Optimization of Parameters} \ \
Because the marginal distribution $p_{\theta}(X, \Yn)$ is usually intractable, to learn the set of parameters $\{\theta_1,\theta_2,\phi_1, \phi_2\}$ given only noisy data, we follow the variational inference framework \citep{blei2017variational} to minimize the negative evidence lower-bound $-\mathrm{ELBO}(x, \yn)$ of the marginal likelihood of each datapoint $(x, \yn)$ instead of maximizing the marginal likelihood itself. By ensembling our decoder and encoder networks, $-\mathrm{ELBO}(x, \yn)$ is derived as follows: 
\begin{align}\label{elbo}
    -\mathrm{ELBO}(x, \yn)
    % &=\mathbb{E}_{(z,y)\sim q_{\phi}(Z,Y|x)}\left[ \log \px \right] +\mathbb{E}_{y\sim q_{\phi_1}(Y|x)}\left[ \log \pyn \right] + kl(q_{\phi_1}(Y|x) |p(Y)) \nonumber\\
    % & - \mathbb{E}_{y\sim q_{\phi_1}(Y|x)}\left[ kl( q_{\phi}(Z|y,x)\| p(Z)) \right]+ kl( q_{\phi}(z,y|\yn,x)\| p_{\theta_3}(z,y|\yn,x)) \nonumber\\
    &=-\mathbb{E}_{(z,y)\sim q_{\phi}(Z,Y|x)}\left[ \log \px \right] -\mathbb{E}_{y\sim q_{\phi_1}(Y|x)}\left[ \log \pyn \right] \nonumber\\
    &  + kl(q_{\phi_1}(Y|x) \|p(Y)) + \mathbb{E}_{y\sim q_{\phi_1}(Y|x)}\left[ kl( q_{\phi_2}(Z|y,x)\| p(Z)) \right],
\end{align}
where $kl(\cdot)$ is the Kullback–Leibler divergence between two distributions. The derivation details is left out in Appendix~A.
Our model learns the class-conditional distribution $P(X|Y)$ by maximizing the first expectation in ELBO, which is equivalent to minimize the reconstruction loss \citep{kingma2013auto}. 
By learning $P(X)$, the inference network $\Qy$ has to select a suitable parameter $\phi^*$ which samples the $y$ and $z$ to minimize the reconstruction loss $\mathbb{E}_{(z,y)\sim q_{\phi}(Z,Y|x)}\left[ \log \px \right]$. When the dimension of $Z$ is chosen to be much smaller than the dimension of $X$, to obtain a smaller reconstruction error, the decoder has to utilize the information provided by $Y$, and force the value of $Y$ to be useful for prediction. Furthermore, we constrain the $Y$ to be a one-hot vector, then $Y$ could be a cluster id of which manifold of the $X$ belongs.

So far, the latent variable $Y$ can be inferred as a cluster id instead of a clean class id. To further link the clusters to clean labels, a naive approach is to select some reliable examples and keep the cluster numbers to be consistent with the noisy labels on these examples. In such a way, the latent representation $Z$ and clean label $Y$ can be effectively inferred, and therefore, it encourages the identifiability of the transition relationship $\Pyn$.
To achieve this, instead of explicitly selecting the reliable example in advance, our method is trained in an end-to-end favor, i.e., the reliable examples are selected dynamically during the update of parameters of our model by using the co-teaching technique \citep{han2018co}. The advantage of this approach is that the selection bias of the reliable example \citep{cheng2020learning} can be greatly reduced. Intuitively, the accurately selected reliable examples can encourage the identifiability of $\Pyn$ and $\Px$, and the accurately estimated $\Pyn$ and $\Px$ will encourage the network to select more reliable examples.
\begin{algorithm}[t]
\caption{CausalNL}\label{pseudocode}
\begin{algorithmic}
\STATE {\textbf{Input:} 
A noisy sample $S$,
Average noise rate $\rho$,
Total epoch $T_{max}$,
Batch size $N$
}.
\end{algorithmic}
\begin{algorithmic}[1]
\STATE \textbf{For} T = $1, \dots, T_{max}$:
\STATE \ \ \textbf{For} mini-batch $\bar{S}=\{x_i\}_{i=0}^N, \tilde{L}=\{\tilde{y}_i\}_{i=0}^N$ in $S$:
\STATE \ \  \ \   Feed $\bar{S}$ to encoders $\hat{q}_{\phi_1^1}$ and $\hat{q}_{\phi_1^2}$ to get clean label sets $L_1$ and $L_2$, respectively;
\STATE \ \  \ \  Feed $(\bar{S}, L_1)$ to encoder $\hat{q}_{\phi_2^1}$ to get a representation set $H_1$, feed $(\bar{S}, L_2)$ to $\hat{q}_{\phi_2^2}$ to get $H_2$;
\STATE \ \  \ \   Update $\hat{q}_{\phi_2^1}$ and $\hat{q}_{\phi_2^2}$ with co-teaching loss;
\STATE \ \  \ \   Feed $(L_1, H_1)$ to decoder $\hat{p}_{\theta_1^1}$ to get reconstructed dataset $\bar{S}_1$, feed $(L_2, H_2)$ to $\hat{p}_{\theta_1^2}$ to get $\bar{S}_2$;
\STATE \ \  \ \  Feed $(\bar{S}_1, L_1)$ to decoder $\hat{p}_{\theta_2^1}$ to get predicted noisy labels $\tilde{L}_1$, feed $(\bar{S}_2, L_2)$ to $\hat{p}_{\theta_2^2}$ to get $\tilde{L}_2$;
\STATE \ \  \ \  Update networks $\hat{q}_{\phi_1^1}$, $\hat{q}_{\phi_2^1}$, $\hat{p}_{\theta_1^1}$ and $\hat{p}_{\theta_2^1}$ by calculating ELBO on $(\bar{S}, \bar{S}_1, \tilde{L}, \tilde{L}_1)$, update networks $\hat{q}_{\phi_1^2}$, $\hat{q}_{\phi_2^2}$, $\hat{p}_{\theta_1^2}$ and $\hat{p}_{\theta_2^2}$ by calculating ELBO on $(\bar{S}, \bar{S}_2, \tilde{L}, \tilde{L}_2)$;
\end{algorithmic}
\begin{algorithmic}
\STATE {\textbf{Output:}
The inference network $\hat{q}_{\phi_1^1}$.}
\end{algorithmic}
\label{alg:1}
\end{algorithm}
\subsection{Practical Implementation} 
Our method is summarized in Algorithm \ref{alg:1} and illustrated in Fig.~\ref{fig:wf}, Here we introduce the structure of our model and loss functions.

\textbf{Model Structure} \ \ 
Because we incorporate co-teaching in our model training, we need to add a copy of the decoders and encoders in our method. As the two branches share the same architectures, we first present the details of the first branch and then briefly introduce the second branch.

For the first branch, we need a set of encoders and decoders to model the distributions in Eq.~\eqref{decoder} and Eq.~\eqref{encoder}. Specifically, we have two encoder networks 
\begin{align*}
    & Y_1 = \hat{q}_{\phi_1^1}(X), \ \ Z_1 \sim \hat{q}_{\phi_2^1}(X, Y_1)
\end{align*}
for Eq.~\eqref{encoder} and two decoder networks
\begin{align*}
    & X_1 = \hat{p}_{\theta_1^1}(Y_1, Z_1), \ \ \tilde{Y}_1 = \hat{p}_{\theta_2^1}(X_1, Y_1)
\end{align*}
for Eq.~\eqref{decoder}.
% Similarly, for Eq.~\eqref{decoder}, the structural equation model is 
% \begin{align*}
%     & X_1 = \hat{p}_{\theta_1^1}(Y_1, Z_1), \ \ \tilde{Y}_1 = \hat{p}_{\theta_2^1}(X_1, Y_1).
% \end{align*}
% Specifically, we have two encoder networks and two decoder networks. 
The first encoder $\hat{q}_{\phi_1^1}(X)$ takes an instance $X$ as input $\hat{q}_{\phi_1^1}(X)$ and output a predicted clean label $Y_1$. The second encoder $\hat{q}_{\phi_2^1}(X, Y_1)$ takes both the instance $X$ and the generated label $Y_1$ as input and outputs a latent feature $Z_1$. Then the generated $Y_1$ and $Z_1$ are passed to the decoder $\hat{p}_{\theta_1^1}(Y_1, Z_1)$ which will generate a reconstructed image $X_1$. Finally, the generated $X_1$ and $Y_1$ will be the input for another decoder  $\hat{p}_{\theta_2^1}(X_1, Y_1)$ which returns predicted noisy labels $\tilde{Y}_1$. It is worth to mention that the reparameterization trick \citep{kingma2013auto} is used for sampling, so as to allow backpropagation in $\hat{q}_{\phi_2^1}(X, Y_1)$.

Similarly, the encoder and decoder networks in the second branch are defined as follows 
\begin{align*}
    & Y_2 = \hat{q}_{\phi_1^2}(X), \ \ Z_2 \sim \hat{q}_{\phi_2^2}(X, Y_2),
    & X_2 = \hat{p}_{\theta_1^2}(Y_2, Z_2), \ \ \tilde{Y}_2 = \hat{p}_{\theta_2^2}(X_2, Y_2).
\end{align*}
During training, we let two encoders $\hat{q}_{\phi_1^1}(X)$ and $\hat{q}_{\phi_1^2}(X)$ teach each other
given every mini-batch.  

\textbf{Loss Functions} \ \  
We divide the loss functions into two parts. The first part is the negative ELBO in Eq.~\eqref{elbo}, and the second part is a co-teaching loss. The detailed formulation will be leaved in Appendix~B.

For the negative ELBO, the first term $-\mathbb{E}_{(z,y)\sim q_{\phi}(Z,Y|x)}\left[ \log \px \right]$ is a reconstruction loss, and we use the $\ell 1$ loss for reconstruction. The second term is $-\mathbb{E}_{y\sim q_{\phi_1}(Y|x)}\left[ \log \pyn \right]$, which aims to learn noisy labels given inference $y$ and $x$, this can be simply replaced by using cross-entropy loss on outputs of both decoders $\hat{p}_{\theta_2^1}(X_1, Y_1)$ and $\hat{p}_{\theta_2^2}(X_2, Y_2)$ with the noisy labels contained in the training data.
The additional two terms are two regularizers. To calculate $kl(q_{\phi_1}(Y|x) \|p(Y))$, we assume that the prior $P(Y)$ is a uniform distribution. Then minimizing $kl(q_{\phi_1}(Y|x) \|p(Y))$ is equivalent to maximizing the entropy of $q_{\phi_1}(Y|x)$ for each instance $x$, i.e., $-\sum_{y}q_{\phi_1}(y|x) \log q_{\phi_1}(y|x)$. The benefit for having this term is that it could reduce the overfiting problem of the inference network. For $\mathbb{E}_{y\sim q_{\phi_1}(Y|x)}\left[ kl( q_{\phi_2}(Z|y,x)\| p(Z))\right]$, we let $p(Z)$ to be a standard multivariate Gaussian distribution. Since, empirically, $q_{\phi_2}(Z|y,x)$ is the encoders $\hat{q}_{\phi_1^1}(X)$ and $\hat{q}_{\phi_1^2}(X)$, and the two encoders are designed to be deterministic mappings. Therefore, the expectation can be removed, and only the $kl$ term $kl( q_{\phi_2}(Z|y,x)\| p(Z))$ is left. When $p(Z)$ is a Gaussian distribution, the kl term could have a closed form solution \citep{kingma2013auto}, i.e., $-\frac{1}{2}\sum_{j=1}^{J}(1+\log((\sigma_j)^2)-(\mu_j)^2-(\sigma_j)^2)$, where $J$ is the dimension of a latent representation $z$, $\sigma_j$ and $\mu_j$ are the encoder outputs.

For the co-teaching loss, we follow the work of Han et al.~\citep{han2018co}. Intuitively, two encoders $\hat{q}_{\phi_1^1}(X)$ and $\hat{q}_{\phi_1^2}(X)$ feed forward all data and selects some data of possibly clean labels. Then, two networks communicate with each other to select possible clean data in this mini-batch and use them for training. Finally, each encoder backpropagates over the data selected by its peer network and updates itself by cross-entropy loss.

\section{Experiments} \label{section4}
In this section, we compare the classification accuracy of proposed method with the popular label-noise learning algorithms \citep{liu2016classification,patrini2017making,jiang2018mentornet,han2018co,xia2019anchor, zhang2017mixup,malach2017decoupling} on both synthetic and real-world datasets. 

\subsection{Experimental Setup}
\textbf{Datasets} \ \ 
We verify the efficacy of our approach on the manually corrupted version of four datasets, i.e., \textit{FashionMNIST} \citep{xiao2017fashion}, \textit{SVHN} \citep{netzer2011svhn}, \textit{CIFAR10}, \textit{CIFAR100} \citep{krizhevsky2009learning}, and one real-world noisy dataset, i.e., \textit{Clothing1M} \citep{xiao2015learning}. \textit{FashionMNIST} contains 60,000 training images and 10,000 test images with 10 classes; \textit{SVHN} contains 73,257 training images and 26,032 test images with 10 classes. \textit{CIFAR10} contains 50,000 training images and 10,000 test images.
\textit{CIFAR10} and \textit{CIFAR100} both contain 50,000 training images and 10,000 test images but the former have 10 classes of images, and later have 10 classes of images.
The four datasets contain clean data. We add instance-dependent label noise to the training sets manually according to Xia et al.~\citep{xia2020part}. \textit{Clothing1M} has 1M images with real-world noisy labels and 10k images with clean labels for testing. For all the synthetic noisy datasets, the experiments have been repeated for 5 times.

\textbf{Network structure and optimization} \ \ 
For fair comparison, all experiments are conducted on NVIDIA Tesla V100, and all methods are implemented by PyTorch. 
Dimension of the latent representation $Z$ is set to 25 for all synthetic noisy datasets.
For encoder networks $\hat{q}_{\phi_1^1}(X)$ and $\hat{q}_{\phi_1^2}(X)$, we use the same network structures with baseline method. Specially, we use a ResNet-18 network for \textit{FashionMNIST}, a ResNet-34 network for \textit{SVHN} and \textit{CIFAR10}. a ResNet-50 network for \textit{CIFAR100} without pretraining. For \textit{Clothing1M}, we use ResNet-50 networks pre-trained on ImageNet. We use random-crop and horizontal flip for data augmentation.
For Clothing1M, we use a ResNet-50 network pre-trained on ImageNet, and the clean training data is not used. Dimension of the latent representation $Z$ is set to 100. Due to limited space, we leave the detailed structure of other decoders and encoders in Appendix~C.

\textbf{Baselines and measurements} \ \ We compare the proposed method with the following state-of-the-art approaches: (i). CE, which trains the standard deep network with the cross entropy loss on noisy datasets. (ii). Decoupling \citep{malach2017decoupling}, which trains two networks on samples whose the predictions from the two networks are different. (iii). MentorNet \citep{jiang2018mentornet}, Co-teaching \citep{han2018co}, which mainly handles noisy labels by training on instances with small loss values. (iv). Forward \citep{patrini2017making}, Reweight \citep{liu2016classification}, and T-Revision \citep{xia2019anchor}. These approaches utilize a class-dependent transition matrix $T$ to correct the loss function. We report average test auracy on over the last ten epochs of each model on the clean test set. Higher classification accuracy means that the algorithm is more robust to the label noise.

\begin{table}[!t]
	\centering
	\caption{Means and standard deviations (percentage) of classification accuracy on \textit{FashionMNIST} with different label noise levels. }
	\label{tab:1}
	{
		\begin{tabular}{cccccc}
		%\hline\cline{1-5}
           	\toprule
			 ~& IDN-20\%      & IDN-30\%      & IDN-40\%    & IDN-45\%  & IDN-50\%         \\ \midrule
			 CE & 88.54$\pm$0.32 & 88.38$\pm$0.42 & 84.22$\pm$0.35 & 69.72$\pm$0.72 & 52.32$\pm$0.68\\
			 
             Co-teaching & 91.21$\pm$0.31 & 90.30$\pm$0.42 & 89.10$\pm$0.29  & 86.78$\pm$0.90 & 63.22$\pm$1.56 \\			 
             Decoupling  & 90.70$\pm$0.28 & 90.34$\pm$0.36 & 88.78$\pm$0.44 & 87.54$\pm$0.53 & 68.32$\pm$1.77 \\ 
		     MentorNet & 91.57$\pm$0.29 & 90.52$\pm$0.41 & 88.14$\pm$0.76 & 85.12$\pm$0.76 & 61.62$\pm$1.42 \\
	 
			 Mixup & 88.68$\pm$0.37 & 88.02$\pm$0.37 & 85.47$\pm$0.55  & 79.57$\pm$0.75 & 66.02$\pm$2.58 \\

			 Forward & 90.05$\pm$0.43 & 88.65$\pm$0.43 & 86.27$\pm$0.48  & 73.35$\pm$1.03 & 58.23$\pm$3.14 \\
			 Reweight & 90.27$\pm$0.27 & 89.58$\pm$0.37 & 87.04$\pm$0.32  & 80.69$\pm$0.89 & 64.13$\pm$1.23 \\
			 T-Revision & \textbf{91.58}$\pm$0.31 & 90.11$\pm$0.61 & 89.46$\pm$0.42  & 84.01$\pm$1.14 & 68.99$\pm$1.04 \\
			 \midrule
			 CausalNL & 90.84$\pm$0.31 & \textbf{90.68}$\pm$0.37 & \textbf{90.01}$\pm$0.45  & \textbf{88.75}$\pm$0.81 & \textbf{78.19}$\pm$1.01 \\
			 
             \bottomrule
		\end{tabular}
	}
\end{table}

\begin{table}[!t]
	\centering
	\caption{Means and standard deviations (percentage) of classification accuracy on \textit{SVHN} with different label noise levels. }
	\label{tab:2}
	{
		\begin{tabular}{cccccc}
		%\hline\cline{1-5}
           	\toprule
           			 ~& IDN-20\%      & IDN-30\%      & IDN-40\%    & IDN-45\%  & IDN-50\%         \\ \midrule
			 CE & 91.51$\pm$0.45 & 91.21$\pm$0.43 & 87.87$\pm$1.12 & 67.15$\pm$1.65 & 51.01$\pm$3.62\\
			 
			 Co-teaching & 93.93$\pm$0.31 & 92.06$\pm$0.31 & 91.93$\pm$0.81  & 89.33$\pm$0.71 & 67.62$\pm$1.99 \\
             Decoupling  & 90.02$\pm$0.25 & 91.59$\pm$0.25 & 88.27$\pm$0.42 & 84.57$\pm$0.89 & 65.14$\pm$2.79 \\ 
		     MentorNet & \textbf{94.08}$\pm$0.12 & 92.73$\pm$0.37 & 90.41$\pm$0.49 & 87.45$\pm$0.75 & 61.23$\pm$2.82\\
		     
			 Mixup & 89.73$\pm$0.37 & 90.02$\pm$0.35 & 85.47$\pm$0.55  & 82.41$\pm$0.62 & 68.95$\pm$2.58 \\

			 Forward & 91.89$\pm$0.31 & 91.59$\pm$0.23 & 89.33$\pm$0.53  & 80.15$\pm$1.91 & 62.53$\pm$3.35 \\
			 Reweight & 92.44$\pm$0.34 & 92.32$\pm$0.51 & 91.31$\pm$0.67  & 85.93$\pm$0.84 & 64.13$\pm$3.75 \\
			 T-Revision & 93.14$\pm$0.53 & 93.51$\pm$0.74 & 92.65$\pm$0.76  & 88.54$\pm$1.58 & 64.51$\pm$3.42 \\
			 \midrule
			 CausalNL & 94.06$\pm$0.23 & \textbf{93.86}$\pm$0.37 & \textbf{93.82}$\pm$0.45  & \textbf{93.19}$\pm$0.81 & \textbf{85.41}$\pm$2.95 \\
             \bottomrule
		\end{tabular}
	}
\end{table}
\begin{table}[!t]
	\centering
	\caption{Means and standard deviations (percentage) of classification accuracy on \textit{CIFAR10} with different label noise levels. }
	\label{tab:3}
	{
		\begin{tabular}{cccccc}
		%\hline\cline{1-5}
           	\toprule
			 ~& IDN-20\%      & IDN-30\%      & IDN-40\%    & IDN-45\%  & IDN-50\%         \\ \midrule
			 CE & 75.81$\pm$0.26 & 69.15$\pm$0.65 & 62.45$\pm$0.86 & 51.72$\pm$1.34 & 39.42$\pm$2.52\\
			 
			 Co-teaching & 80.96$\pm$0.31 & 78.56$\pm$0.61 & 73.41$\pm$0.78  & 71.60$\pm$0.79 & 45.92$\pm$2.21 \\
             Decoupling  & 78.71$\pm$0.15 & 75.17$\pm$0.58 & 61.73$\pm$0.34 & 58.61$\pm$1.73 & 50.43$\pm$2.19 \\ 
		     MentorNet & 81.03$\pm$0.12 & 77.22$\pm$0.47 & 71.83$\pm$0.49 & 66.18$\pm$0.64 & 47.89$\pm$2.03 \\
		     
			 Mixup & 73.17$\pm$0.37 & 70.02$\pm$0.31 & 61.56$\pm$0.71  & 56.45$\pm$0.62 & 48.95$\pm$2.58 \\

			 Forward & 74.64$\pm$0.32 & 69.75$\pm$0.56 & 60.21$\pm$0.75  & 48.81$\pm$2.59 & 46.27$\pm$1.30 \\
			 Reweight & 76.23$\pm$0.25 & 70.12$\pm$0.72 & 62.58$\pm$0.46  & 51.54$\pm$0.92 & 45.46$\pm$2.56 \\
			 T-Revision& 76.15$\pm$0.37 & 70.36$\pm$0.61 & 64.09$\pm$0.37  & 52.42$\pm$1.01 & 49.02$\pm$2.13 \\
			 \midrule
			 CausalNL & \textbf{81.47}$\pm$0.32 & \textbf{80.38}$\pm$0.37 & \textbf{77.53}$\pm$0.45  & \textbf{78.60}$\pm$0.93 & \textbf{77.39}$\pm$1.24 \\
			 \bottomrule
		\end{tabular}
	}
\end{table}

\begin{table}[!t]
	\centering
	\caption{Means and standard deviations (percentage) of classification accuracy on \textit{CIFAR100} with different label noise levels. }
	\label{tab:4}
	{
		\begin{tabular}{cccccc}
		%\hline\cline{1-5}
           	\toprule
			 ~& IDN-20\%      & IDN-30\%      & IDN-40\%    & IDN-45\%  & IDN-50\%         \\ \midrule
			 CE & 30.42$\pm$0.44 & 24.15$\pm$0.78 & 21.45$\pm$0.70 & 15.23$\pm$1.32 & 14.42$\pm$2.21\\
			 
			 Co-teaching & 37.96$\pm$0.53 & 33.43$\pm$0.74 & 28.04$\pm$1.43  & 25.60$\pm$0.93 & 23.97$\pm$1.91 \\
             Decoupling  & 36.53$\pm$0.49 & 30.93$\pm$0.88 & 27.85$\pm$0.91 & 23.81$\pm$1.31 & 19.59$\pm$2.12 \\ 
		     MentorNet & 38.91$\pm$0.54 & 34.23$\pm$0.73 & 31.89$\pm$1.19 & 27.53$\pm$1.23 & 24.15$\pm$2.31 \\
		     
			 Mixup & 32.92$\pm$0.76 & 29.76$\pm$0.87 & 25.92$\pm$1.26  & 23.13$\pm$2.15 & 21.31$\pm$1.32 \\

			 Forward & 36.38$\pm$0.92 & 33.17$\pm$0.73 & 26.75$\pm$0.93  & 21.93$\pm$1.29 & 19.27$\pm$2.11 \\
			 Reweight & 36.73$\pm$0.72 & 31.91$\pm$0.91 & 28.39$\pm$1.46  & 24.12$\pm$1.41 & 20.23$\pm$1.23 \\
			 T-Revision& 37.24$\pm$0.85 & 36.54$\pm$0.79 & 27.23$\pm$1.13  & 25.53$\pm$1.94 & 22.54$\pm$1.95 \\
			 \midrule
			 CausalNL & \textbf{41.47}$\pm$0.32 & \textbf{40.98}$\pm$0.62 & \textbf{34.02}$\pm$0.95  & \textbf{33.34}$\pm$1.13 & \textbf{32.129}$\pm$2.23 \\
			 \bottomrule
		\end{tabular}
	}
\end{table}

\begin{table}[!t]
% \vspace{-10pt}
	\centering
	%\small
	\caption{Classification accuracy on \textit{Clothing1M}. In the experiments, only noisy samples are exploited to train and validate the deep model.}
	\label{tab:5}
	{
		\begin{tabular}{ccccccccc}
		%\hline\cline{1-5}
           	\toprule
			 CE &Decoupling & MentorNet & Co-teaching & Forward & Reweight &  T-Revision &  caualNL\\ \midrule
			 68.88 & 54.53 & 56.79 & 60.15 & 69.91 & 70.40 & 70.97& \textbf{72.24} \\ \toprule
             %\bottomrule
		\end{tabular}
	}
\end{table}

\subsection{Classification accuracy Evaluation}
\textbf{Results on synthetic noisy datasets} \ \
Tables \ref{tab:1}, \ref{tab:2},  \ref{tab:3}, and \ref{tab:4} report the classification accuracy on the datasets of \textit{F-MNIST}, \textit{SVHN}, \textit{CIFAR-10}, and \textit{CIFAR100}, respectively.  
The synthetic experiments reveal that our method is powerful in handling instance-dependent label noise particularly in the situation of high noise rates. For all ddatasets, the classification accuracy does not drop too much compared with all baselines, and the advantages of our proposed method increase with the increasing of the noise rate. Additional, it shows that for all these dataset $Y$ should be a cause of $X$, and therefore, the classification accuracy by using our method can be improved.

For noisy \textit{F-MNIST}, \textit{SVHN} and \textit{CIFAR-10}, in the easy case IDN-20\%, almost all methods work well. When the noise rate is 30\%, the advantages of causalNL begin to show. We surpassed all methods obviously. When the noise rate raises, all the baselines are gradually defeated. Finally, in the hardest case, i.e., IDN-50\%, the superiority of causalNL widens the gap of performance. The classification accuracy of causalNL is at least over 10\% higher than the best baseline method.
For noisy \textit{CIFAR-100}, all the methods do not work well. However, causalNL still overtakes the other methods with clear gaps for all different levels of noise rate.

\textbf{Results on the real-world noisy dataset} \ \ 
On the the real-world noisy dataset \textit{Clothing1M}, our method causalNL outperforms all the baselines as shown in Table~\ref{tab:5}. The experimental results also shows that the noise type in \textit{Clothing1M} is more likely to be instance-dependent label noise, and making the instance-independent assumption on the transition matrix sometimes can be strong.

\section{Conclusion} \label{section5}
In this paper, we have investigated how to use $P(X)$ to help learn instance-dependent label noise. Specifically, the previous assumptions are made on the transition matrix, and the assumptions are hard to be verified and might be violated on real-world datasets. 
Inspired by a causal perspective, when $Y$ is a cause of $X$, then $P(X)$ should contain useful information to infer the clean label $Y$. We propose a novel generative approach called causalNL for instance-dependent label-noise learning. Our model makes use of the causal graph to contribute to the identifiability of the transition matrix, and therefore help learn clean labels. In order to learn $P(X)$, compared to the previous methods, our method contains more parameters. But the experiments on both synthetic and real-world noisy datasets show that a little bit sacrifice on computational  efficiency is worth, i.e., the classification accuracy of casualNL significantly outperforms all the state-of-the-art methods. Additionally, the results also tells us that in classification problems, $Y$ can usually be considered as a cause of $X$, and suggest that the understanding and modeling of data-generating process can help leverage additional information that is useful in solving advanced machine learning problems concerning the relationship between different modules of the data joint distribution.  
In our future work, we will study the theoretical properties of our method and establish the identifiability result under certain assumptions on the data-generative process.
% \section*{Broader Impact} \label{section5}

\section*{Acknowledgments} 

TL was partially supported by Australian Research Council Projects DP-180103424, DE-190101473, and IC-190100031. 
GM was supported by Australian Research Council Project DE210101624. 
BH was supported by supported by the RGC Early Career Scheme No. 22200720 and NSFC Young Scientists Fund No. 62006202.
GN was supported by JST AIP Acceleration Research Grant Number JPMJCR20U3, Japan.
KZ was supported in part by the National Institutes of Health (NIH) under Contract R01HL159805, by the NSF-Convergence Accelerator Track-D award \#2134901, and by a grant from Apple. The NIH or NSF is not responsible for the views reported in this article.

\bibliographystyle{plainnat}
\bibliography{bib}

\appendix
\section{Derivation Details of evidence lower-bound (ELBO)}\label{appendixA}
In this section, we show the derivation details of $\mathrm{ELBO}(x,\tilde{y})$. \\
Recall that the causal decomposition of the instance-dependent label noise is 
\begin{equation}
    P(X,\tilde{Y}, Y,Z) = P(Y)P(Z)P(X|Y,Z)P(\tilde{Y}|Y,X).
    \label{eq:joint}
\end{equation}
Our encoders model following distributions 
\begin{equation}\label{encoder_o}
    q_{\phi}(Z,Y|X) = \Qz \Qy,
\end{equation}

and decoders model the following distributions
\begin{equation}\label{decoder}
 p_\theta(X, \Yn|Y, Z) = \Px\Pyn.
\end{equation}
Now, we start with maximizing the log-likelihood $p_{\theta}(x, \yn)$ of each datapoint $(x, \yn)$.
\begin{align}\label{elbo}
    \log p_{\theta}(x, \yn)
    &=\log \int_{z}\int_{y} p_{\theta}(x, \yn, z, y)\mathrm{d}y \mathrm{d}z  \nonumber\\
    &=\log \int_{z}\int_{y} p_{\theta}(x, \yn, z, y) \frac{\q}{\q} \mathrm{d}y \mathrm{d}z \nonumber\\
    &=\log \mathbb{E}_{(z,y)\sim q_{\phi}(Z,Y|x)} \left[ \frac{p_{\theta}(x, \yn, z, y)}{\q} \right] \nonumber\\
    &\geq \mathbb{E}_{(z,y)\sim q_{\phi}(Z,Y|x)} \left[ \log \frac{p_{\theta}(x, \yn, z, y)}{\q} \right] := \mathrm{ELBO}(x,\tilde{y})\nonumber\\
    &= \mathbb{E}_{(z,y)\sim q_{\phi}(Z,Y|x)} \left[ \log \frac{ p(z)p(y)\px\pyn)}{\q} \right] \nonumber\\
    &= \mathbb{E}_{(z,y)\sim q_{\phi}(Z,Y|x)} [ \log\left(\px\right)]+ \mathbb{E}_{(z,y)\sim q_{\phi}(Z,Y|x)} [\log\left(\pyn \right)] \nonumber\\
    &+\mathbb{E}_{(z,y)\sim q_{\phi}(Z,Y|x)} \left[\log\left(\frac{p(z)p(y)}{\qz \qy}\right)\right] 
\end{align}
The $\mathrm{ELBO}(x,\tilde{y})$ above can be further simplified. Specifically, 
\begin{align}\label{term2}
&\mathbb{E}_{(z,y)\sim q_{\phi}(Z,Y|x)}[\log\left(\pyn \right)]
= \mathbb{E}_{y\sim q_{\phi_1}(Y|x)} \mathbb{E}_{z\sim q_{\phi_2}(Z|y,x)}[\log\left(\pyn \right)] \nonumber\\
=& \mathbb{E}_{y\sim q_{\phi_1}(Y|x)}  [\log\left(\pyn \right)],
\end{align}
and similarly,
\begin{align} \label{term3}
&\mathbb{E}_{(z,y)\sim q_{\phi}(Z,Y|x)} \left[\log\left(\frac{p(z)p(y)}{\qz \qy}\right)\right]   \nonumber\\
=&\mathbb{E}_{y\sim q_{\phi_1}(Y|x)} \mathbb{E}_{z\sim q_{\phi_2}(Z|y,x)} \left[\log\left(\frac{p(z)p(y)}{\qz \qy}\right)\right]  \nonumber\\
=&\mathbb{E}_{y\sim q_{\phi_1}(Y|x)} \mathbb{E}_{z\sim q_{\phi_2}(Z|y,x)} \left[\log\left(\frac{p(y)}{ \qy}\right)\right] +\mathbb{E}_{y\sim q_{\phi_1}(Y|x)} \mathbb{E}_{z\sim q_{\phi_2}(Z|y,x)} \left[\log\left(\frac{p(z)}{ \qz}\right)\right]  \nonumber\\
=&\mathbb{E}_{y\sim q_{\phi_1}(Y|x)}  \left[\log\left(\frac{p(y)}{ \qy}\right)\right] +\mathbb{E}_{y\sim q_{\phi_1}(Y|x)} \mathbb{E}_{z\sim q_{\phi_2}(Z|y,x)} \left[\log\left(\frac{p(z)}{ \qz}\right)\right]  \nonumber\\
=& -kl(q_{\phi_1}(Y|x) \|p(Y)) -\mathbb{E}_{y\sim q_{\phi_1}(Y|x)}\left[ kl( q_{\phi_2}(Z|y,x)\| p(Z)) \right],
\end{align}

By combing Eq.~\eqref{elbo}, Eq.~\eqref{term2} and  Eq.~\eqref{term3}, we get
\begin{align}
    \mathrm{ELBO}(x, \yn)
    &=\mathbb{E}_{(z,y)\sim q_{\phi}(Z,Y|x)}\left[ \log \px \right] +\mathbb{E}_{y\sim q_{\phi_1}(Y|x)}\left[ \log \pyn \right] \nonumber\\
    &  - kl(q_{\phi_1}(Y|x) \|p(Y)) - \mathbb{E}_{y\sim q_{\phi_1}(Y|x)}\left[ kl( q_{\phi_2}(Z|y,x)\| p(Z)) \right],
\end{align}
which is the ELBO in our main paper.

\section{Loss Functions}\label{appendixB}
In this section, we provide the empirical solution of the ELBO and co-teaching loss.
Remind that our encoder networks and decoder networks in the the first branch are defined as follows 
\begin{align*}
    & Y_1 = \hat{q}_{\phi_1^1}(X), \ \ Z_1 \sim \hat{q}_{\phi_2^1}(X, Y_1),
    & X_1 = \hat{p}_{\theta_1^1}(Y_1, Z_1), \ \ \tilde{Y}_1 = \hat{p}_{\theta_2^1}(X_1, Y_1),
\end{align*}

Let $S$ be the noisy training set, and $d^2$ be the dimension of an instance $x$. Let $y_1$ and $z_1$ be the estimated clean label and latent representation for the instance $x$, respectively, by the first branch. As mentioned in our main paper (see Section 3.2), the negative ELBO loss is to minimize 
1). a reconstruction loss between each instance $x$ and $\hat{p}_{\theta_1^1}(x,y_1)$; 2). a cross-entropy loss between noisy labels $\hat{p}_{\theta_2^1}(x_1, x_1)$ and $\tilde{y}$;
3). a cross-entropy loss between $\hat{q}_{\phi_2^1}(x, y_1)$ and uniform distribution $P(Y)$;
4). a cross-entropy loss between $\hat{q}_{\phi_2^1}(x, y_1)$ and Gaussian distribution $P(Z)$. Specifically, the empirical version of the ELBO for the first branch is as follows. 
\begin{align}
   \sum_{(x,\tilde{y})\in S}  \hat{\mathrm{ELBO}}^1(x, \yn)
    &=  \sum_{(x,\tilde{y})\in S} \left[ \beta_0 \frac{1}{d^2}\|x-\hat{p}_{\theta_1^1}(y_1, z_1)\|_1
    -\beta_1 \yn \log \hat{p}_{\theta_2^1}(x_1, y_1) 
    +\beta_2 \hat{q}_{\phi_1^1}(x) \log \hat{q}_{\phi_1^1}(x) \right. \nonumber \\ 
    &\left . +\beta_3 \sum_{j=1}^{J}(1+\log((\sigma_j)^2)-(\mu_j)^2-(\sigma_j)^2) \right].
\end{align}
The hyper-parameter $\beta_0$ and $\beta_1$ are set to $0.1$, and $\beta_2$ are set to $1e-5$ because encouraging the distribution to be uniform on a small min-batch (i.e., $128$) could have a large estimation error. The hyper-parameter $\beta_3$ are set to $0.01$. The empirical version of the ELBO for the second branch shares the same settings as the first branch. 

For co-teaching loss, we directly follow Han et al.~\citep{han2018co}. Intuitively, in each mini-batch, both encoders $\hat{q}_{\phi_1^1}(X)$ and $\hat{q}_{\phi_1^2}(X)$ trust small-loss examples, and exchange the examples to each other by a cross-entropy loss. The number of the small-loss instances used for training decays with respect to the training epoch. The experimental settings for co-teaching loss are same as the settings in the original paper \citep{han2018co}.

\section{More experimental settings}\label{appendixC}
In this section, we summarize the network structures for different datasets. The network structure for modeling $\Qy$ and dimension of the latent representation $Z$ have been discussed in our main paper. For optimization method, we use Adam with the default learning rate $1e-3$ in Pytorch. The source code has been included in our supplementary material.

For \textit{FashionMNIST} \citep{xiao2017fashion}, \textit{SVHN} \citep{netzer2011svhn}, \textit{CIFAR10} and \textit{CIFAR100}, we use the same number of hidden layers and feature maps. Specifically, 1). we model $\Qz$ and $\Pyn$ by two 4-hidden-layer convolutional networks, and the corresponding feature maps are 32, 64, 128 and 256; 2). we model $\Px$ by a 4-hidden-layer transposed-convolutional network, and the corresponding feature maps are 256, 128, 64 and 32. We ran 150 epochs for each experiment on these datasets. 

For \textit{Clothing1M} \citep{xiao2015learning}, 1). we model $\Qz$ and $\Pyn$ by two 5-hidden-layer convolutional networks, and the corresponding feature maps are 32, 64, 128, 256, 512; 2). we model $\Px$ by a 5-hidden-layer transposed-convolutional network, and the corresponding feature maps are  512, 256, 128, 64 and 32. We ran 40 epochs on \textit{Clothing1M}.

\end{document}